\documentclass[conference]{IEEEtran}
\IEEEoverridecommandlockouts
\usepackage{cite}
\usepackage{amsmath,amssymb,amsfonts}
\usepackage{graphicx}
\usepackage{textcomp}
\usepackage{xcolor}
\usepackage{booktabs}
\usepackage{array}
\usepackage{url}
\usepackage[hidelinks]{hyperref}
\usepackage{placeins}
\usepackage{multirow}
\def\BibTeX{{\rm B\kern-.05em{\sc i\kern-.025em b}\kern-.08em
    T\kern-.1667em\lower.7ex\hbox{E}\kern-.125emX}}
\newcommand{\DCAL}{DCAL}
\begin{document}

\title{Distribution-Calibrated Candidate Selection for Personalized Text-to-Image Generation}

\author{
\IEEEauthorblockN{Wenyan Xu}
\IEEEauthorblockA{
\textit{School of Computer Science, Guangdong University of Technology}\\
Guangzhou, China\\
3223004777@mail2.gdut.edu.cn}
}

\maketitle

\begin{abstract}
Subject-driven personalized text-to-image systems commonly generate multiple candidates but return a single output, yet the selection stage is rarely evaluated as an explicit system component. Identity-only selection may overemphasize reference similarity at the expense of prompt alignment and output diversity. We propose distribution-calibrated candidate selection (DCAL), a training-free sequential selector for a predefined ordered prompt set that jointly scores calibrated identity, text alignment, and cross-prompt novelty. All selectors operate on identical pre-generated $K=4$ candidate pools for all 30 DreamBooth subjects using a fixed LoRA-adapted generator. Relative to the fixed-seed Random-of-4 control, DCAL yields higher 1-LPIPS, CLIP-I, DINO-I, and CLIP-T for 22, 25, 28, and 22 of the 30 subjects, respectively. Compared with identity-only selection, DCAL improves CLIP-T (0.3264 versus 0.3179), CLIP diversity (0.2444 versus 0.2368), and DINO diversity (0.3286 versus 0.3170), while CLIP-I and DINO-I decrease from 0.8250 to 0.8204 and from 0.7560 to 0.7453. Compared with identity-text scoring, prompt-wise novelty calibration increases CLIP-T from 0.3253 to 0.3264 and CLIP diversity from 0.2366 to 0.2444. We further formalize finite-pool Pareto and perturbation-stability properties and analyze inference complexity.
\end{abstract}


\section{Introduction}
Subject-driven personalization adapts a pretrained text-to-image model to render a specific subject from a few references under novel prompts. Representative approaches optimize token embeddings, fine-tune the diffusion model, or adapt a compact parameter subset~\cite{gal2022textual,ruiz2023dreambooth,kumari2023custom,hu2021lora}. Recent encoder-based and tuning-free systems further improve personalization efficiency and controllability~\cite{wang2024instantid,guo2024pulid,chan2024sag,feng2025personalize}.

\begin{figure}[t]
\centering
\includegraphics[width=0.94\linewidth]{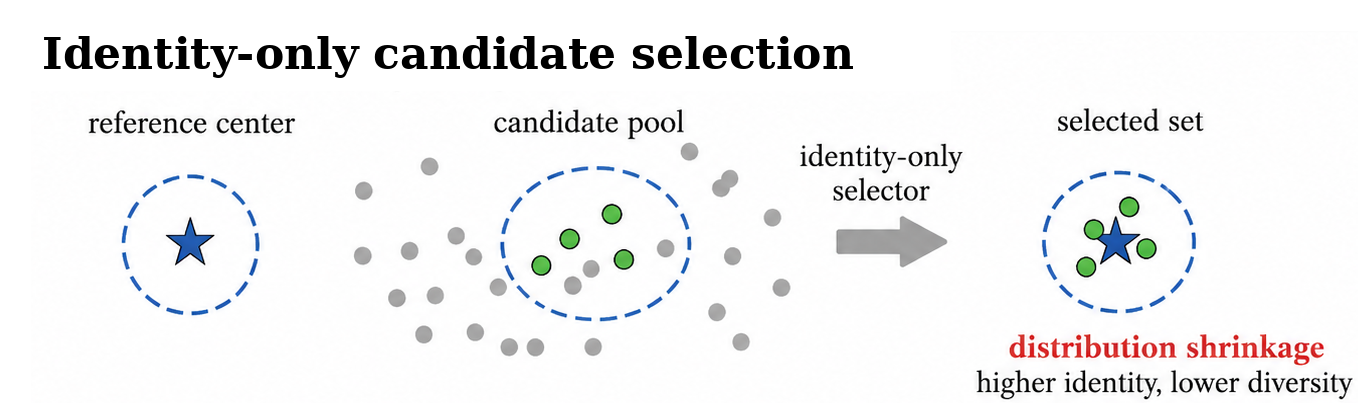}
\caption{Motivation. Identity-only selection concentrates outputs near the reference center; DCAL also considers prompt alignment and cross-prompt novelty relative to earlier selections.}
\label{fig:motivation}
\end{figure}

Most evaluations emphasize the adapted generator, although practical systems often sample several outputs and return only one. This final selection step is not neutral. As illustrated in Fig.~\ref{fig:motivation}, an identity-only selector concentrates results near the reference center; it may improve identity similarity while suppressing prompt-compatible or visually distinct candidates. We therefore study candidate selection as an explicit inference component applied to a fixed LoRA-adapted generator.

Our contributions are threefold. First, we propose \DCAL{}, a training-free sequential selector for a predefined prompt order, with subject-level identity/text calibration and prompt-level novelty calibration. Second, we evaluate random, identity-only, identity-text, identity-diversity, and Pareto-TOPSIS rules on identical pre-generated candidate pools, separating the scoring rule from sampling. Third, we formalize the finite-pool Pareto-efficiency and perturbation-stability conditions and provide an auditable complexity analysis.

\section{Related Work}
\textbf{Personalized generation.} Textual Inversion learns a subject token, DreamBooth fine-tunes the generator with prior preservation, and Custom Diffusion updates restricted attention parameters~\cite{gal2022textual,ruiz2023dreambooth,kumari2023custom}. Compact adaptation methods such as LoRA, SVDiff, and PaRa reduce the optimized parameter space~\cite{hu2021lora,han2023svdiff,chen2024para}. More recent customization methods include InstantID and PuLID for tuning-free identity conditioning, Subject-Agnostic Guidance for balancing subject and prompt conditions, and Personalize Anything for training-free personalization with diffusion transformers~\cite{wang2024instantid,guo2024pulid,chan2024sag,feng2025personalize}. These studies primarily modify conditioning, training, or denoising. Our focus is orthogonal: selecting one result from a fixed candidate pool produced by an already adapted generator.

\textbf{Selection and evaluation.} CLIP and DINOv2 embeddings are widely used to measure text alignment and subject similarity~\cite{radford2021clip,oquab2023dinov2}. A simple test-time strategy selects the candidate with the highest CLIP-T, CLIP-I, DINO-I, or a comprehensive weighted score. Multi-objective sorting and Pareto-style selection preserve non-dominated candidates~\cite{deb2002nsga}; TOPSIS selects an alternative by its distance from ideal and anti-ideal points~\cite{hwang1981topsis}; and maximal marginal relevance (MMR) trades relevance against redundancy~\cite{carbonell1998mmr}. A deployed system must nevertheless specify a deterministic final rule when one image is returned. Learning-to-rank can learn such a rule from preferences~\cite{liu2009learning}, but requires additional supervision or proxy labels. \DCAL{} instead targets an auditable, training-free setting in which identity, text alignment, and cross-prompt redundancy are evaluated over fixed candidate pools under a same-pool protocol.

\section{Method}
\subsection{Problem Setup and Candidate Pool}
Let subject $s$ have reference images $\mathcal{X}_s=\{x_i^s\}$ and a predefined ordered sequence of $P$ target prompts. A fixed DreamBooth LoRA generator produces $K$ candidates $\mathcal{Y}_{s,p}=\{y_{p,k}\}_{k=1}^{K}$ for prompt $p$. The selector returns one index $k_p^*$ for every prompt in sequence without changing the generator or retraining the adapter. In all selector experiments, $P=5$, $K=4$, and every control operates on the same recorded candidate identifiers and predefined benchmark prompt order.

\subsection{Distribution-Calibrated Selection}
Let $\phi_C$ and $\phi_D$ denote the CLIP and DINOv2 image encoders, and let $\psi_C$ denote the CLIP text encoder~\cite{radford2021clip,oquab2023dinov2}. The implementation uses maximum similarity to the reference set:
\begin{align}
c_{p,k}&=\max_i\operatorname{sim}(\phi_C(y_{p,k}),\phi_C(x_i^s)), \nonumber\\
d_{p,k}&=\max_i\operatorname{sim}(\phi_D(y_{p,k}),\phi_D(x_i^s)), \nonumber\\
t_{p,k}&=\operatorname{sim}(\phi_C(y_{p,k}),\psi_C(p)).
\label{eq:rawscores}
\end{align}
For a finite score vector indexed by $\mathcal{A}$, with mean $\mu_z$ and sample standard deviation $\sigma_z$, the implemented transform is
\begin{equation}
N_{\mathcal{A}}(z_a)=
\begin{cases}
(z_a-\mu_z)/\sigma_z, & \epsilon\leq\sigma_z<\infty,\\
z_a-\mu_z, & \text{otherwise},
\end{cases}\quad \epsilon=10^{-6}.
\label{eq:standardize}
\end{equation}
Thus, a zero-variance coordinate becomes identically zero. Identity and text are standardized over the subject's full $PK$ pool:
\begin{equation}
I_{p,k}=\tfrac{1}{2}N_{PK}(c_{p,k})+\tfrac{1}{2}N_{PK}(d_{p,k}),\qquad
T_{p,k}=N_{PK}(t_{p,k}).
\label{eq:calibrated}
\end{equation}
DCAL processes prompts in the predefined benchmark order. Given the previously selected outputs, the raw cross-prompt novelty and its prompt-wise calibrated coordinate are
\begin{equation}
\begin{aligned}
R_{p,k}&=
\begin{cases}
0, & p=1,\\
1-\max_{q<p}\operatorname{sim}\!\left(\phi_C(y_{p,k}),\phi_C(y_{q,k_q^*})\right), & p>1.
\end{cases}\\
D_{p,k}&=N_{K,p}(R_{p,k}).
\end{aligned}
\label{eq:novelty}
\end{equation}
where $N_{K,p}$ applies Eq.~\eqref{eq:standardize} to the $K$ novelty values of prompt $p$. This second calibration is necessary because raw cosine novelty has a much smaller empirical scale than the standardized identity and text coordinates. DCAL is order-dependent because the choice at prompt $p$ depends on earlier selected outputs. For $p=1$, and for the single-prompt case $P=1$, $D_{p,k}=0$ for every candidate, so DCAL reduces exactly to identity-text comprehensive scoring.
The selected index is
\begin{equation}
k_p^*=\arg\max_k\left[\lambda_I I_{p,k}+\lambda_T T_{p,k}+\lambda_D D_{p,k}\right].
\label{eq:dcal}
\end{equation}
We set $\lambda_I=1.00$ and $\lambda_T=0.35$ before the novelty-weight search; neither value was optimized on the 30-subject benchmark. Backpack, cat, and dog2 constitute a fixed three-subject development subset. Holding $\lambda_I$ and $\lambda_T$ constant, we evaluated $\lambda_D\in\{0,0.05,0.10,0.15,0.25,0.35,0.50,0.75,1.00\}$ and selected $\lambda_D=0.15$, the largest tested value for which aggregate CLIP-I, DINO-I, and CLIP-T each retained at least 99\% of the corresponding score obtained with $\lambda_D=0$. The selected weights were then fixed for the remaining 27 subjects and the aggregate evaluation over all 30 subjects; no subject-specific tuning was performed.

Because $\lambda_I,\lambda_T,\lambda_D>0$, at each prompt and conditional on the fixed previously selected set, any maximizer of Eq.~\eqref{eq:dcal} is a supported Pareto-efficient point in the finite normalized objective set formed by $(I_{p,k},T_{p,k},D_{p,k})$. Let $\gamma$ denote the positive gap between the largest and second-largest scalarized scores. If the absolute perturbation of every candidate's scalarized score is smaller than $\gamma/2$, the selected index remains unchanged. These properties concern only the fixed finite pool at one sequential decision step and do not imply global Pareto optimality over the generator distribution.

\subsection{Same-Pool Control Selectors}
Random-of-4 is a fixed-seed reproducible control: it performs one deterministic seeded uniform draw per prompt using seed string ``123:subject''. It is not an estimate of expected random-selection performance. Identity-only-of-4 selects the candidate maximizing the absolute score $0.5c_{p,k}+0.5d_{p,k}$. Because these controls and \DCAL{} reuse exactly the same four images, differences among them are attributable to selection rather than to a different generator or sampling run.

\subsection{Relation to Other Ranking Rules}
Table~\ref{tab:selector_relation} summarizes how \DCAL{} differs from common candidate selectors. The distinction is not merely the use of a weighted score. DCAL standardizes identity and text over the fixed subject-level candidate pool and calibrates sequential novelty over the current prompt pool. If $\lambda_D=0$, Eq.~\eqref{eq:dcal} reduces to identity-text comprehensive scoring. If $\lambda_T=\lambda_D=0$, it retains only the standardized identity coordinate, whereas the reported identity-only control maximizes the absolute identity score. Pareto + TOPSIS uses the same normalized $(I_{p,k},T_{p,k},D_{p,k})$ coordinates and $(\lambda_I,\lambda_T,\lambda_D)$ weights as Eq.~\eqref{eq:dcal}; at each sequential step it retains the non-dominated candidates and selects the largest TOPSIS closeness coefficient within the current prompt pool. MMR balances relevance and redundancy without the same two-level calibration. DCAL keeps every decision traceable to recorded candidate identifiers and component scores.

\begin{table}[t]
\centering
\caption{Relation between DCAL and common candidate selection rules.}
\label{tab:selector_relation}
\scriptsize
\setlength{\tabcolsep}{2.4pt}
\resizebox{\linewidth}{!}{%
\begin{tabular}{lccp{4.1cm}}
\toprule
Selector & Training-free & Same pool & Difference from DCAL \\
\midrule
Random-of-$K$ & Yes & Yes & Does not use identity or text scores. \\
Identity-only & Yes & Yes & Prioritizes reference-nearest outputs. \\
Identity + text & Yes & Yes & Uses Eq.~\eqref{eq:dcal} with $\lambda_D=0$. \\
Identity + diversity & Yes & Yes & Uses Eq.~\eqref{eq:dcal} with $\lambda_T=0$. \\
Pareto + TOPSIS & Yes & Yes & Uses ideal-point tie-breaking on the Pareto set. \\
MMR & Yes & Yes & Trades relevance against redundancy without pool calibration. \\
Learning-to-rank & No & Possible & Requires ranking labels or proxy supervision. \\
\DCAL{} & Yes & Yes & Calibrates identity/text and prompt-wise novelty. \\
\bottomrule
\end{tabular}}
\end{table}

\section{Experiments}
\subsection{Protocol}
Experiments use SDXL and the public DreamBooth 30-subject protocol. The LoRA baseline uses rank $16$, $\alpha=16$, 200 training steps per subject, learning rate $10^{-4}$, $1024\times1024$ resolution, 50 DDIM steps, guidance scale $7.5$, training seed $42$, and sampling seed $0$. CLIP uses ViT-B/32, while the DINO metrics use the DINOv2 ViT-B/14 model \texttt{vit\_base\_patch14\_dinov2.lvd142m}. We report 1-LPIPS, CLIP-I, DINO-I, CLIP-T, CLIP diversity (C-Div), DINO diversity (D-Div), and Pair-LPIPS. Higher is better for the displayed metrics, while identity and diversity represent different objectives.

\begin{table}[t]
\centering
\caption{Results on all 30 DreamBooth subjects. All $K=4$ selectors reuse identical candidate pools. Bold denotes the best result among same-pool $K=4$ selectors.}
\label{tab:main}
\scriptsize
\setlength{\tabcolsep}{2.25pt}
\resizebox{\linewidth}{!}{%
\begin{tabular}{lccccccc}
\toprule
Method & $K$ & 1-LP & C-I & D-I & C-T & C-Div & D-Div \\
\midrule
LoRA r16 & 1 & 0.3585 & 0.8011 & 0.6704 & 0.3219 & 0.2669 & 0.4445 \\
Random-of-4 & 4 & 0.3650 & 0.7974 & 0.6710 & 0.3201 & \textbf{0.2677} & \textbf{0.4307} \\
Identity-only-of-4 & 4 & \textbf{0.3789} & 0.8250 & \textbf{0.7560} & 0.3179 & 0.2368 & 0.3170 \\
Identity + text & 4 & 0.3775 & 0.8246 & 0.7450 & 0.3253 & 0.2366 & 0.3279 \\
Identity + diversity & 4 & 0.3781 & \textbf{0.8264} & 0.7498 & 0.3175 & 0.2376 & 0.3281 \\
Pareto + TOPSIS & 4 & 0.3740 & 0.8244 & 0.7480 & 0.3232 & 0.2383 & 0.3245 \\
\DCAL{} & 4 & 0.3762 & 0.8204 & 0.7453 & \textbf{0.3264} & 0.2444 & 0.3286 \\
\bottomrule
\end{tabular}}
\end{table}

\subsection{Same-Pool Controls}
Table~\ref{tab:main} shows that the gains are not explained by candidate count alone. Relative to the fixed-seed Random-of-4 control, \DCAL{} improves 1-LPIPS from 0.3650 to 0.3762, CLIP-I from 0.7974 to 0.8204, DINO-I from 0.6710 to 0.7453, and CLIP-T from 0.3201 to 0.3264. It achieves higher subject-level values for 22, 25, 28, and 22 of the 30 subjects, respectively, indicating that the improvements are distributed across subjects rather than concentrated in a small subset.

Identity-only-of-4 defines the identity-maximizing endpoint. Compared with it, \DCAL{} increases CLIP-T from 0.3179 to 0.3264, C-Div from 0.2368 to 0.2444, DINO-Div from 0.3170 to 0.3286, and Pair-LPIPS from 0.6441 to 0.6475, while CLIP-I and DINO-I decrease from 0.8250 to 0.8204 and from 0.7560 to 0.7453. These results quantify the trade-off among identity preservation, prompt alignment, and output diversity. On the 27 subjects excluded from weight selection, \DCAL{} retains the same direction relative to identity-text scoring: C-T is 0.3254 versus 0.3243, C-Div is 0.2499 versus 0.2438, and D-Div is 0.3348 versus 0.3342.

The numerical ranker controls isolate the contribution of each coordinate. Relative to identity-text scoring, calibrated novelty increases C-T from 0.3253 to 0.3264, C-Div from 0.2366 to 0.2444, and D-Div from 0.3279 to 0.3286, while C-I decreases from 0.8246 to 0.8204. Identity-diversity attains the highest C-I, but its C-T is 0.3175. Pareto-TOPSIS preserves higher identity than \DCAL{} but produces lower C-T and embedding diversity. No single selector dominates every objective.

\subsection{Qualitative Results}
Figure~\ref{fig:qual} presents qualitative comparisons under identical candidate pools. For both the backpack and cat prompts, random, identity-only, and \DCAL{} return different candidates even though the adapted generator and available images are fixed. These examples illustrate the effect of the selection objective.

\subsection{Selector Audit and Complexity}
The recorded candidate identifiers separate generation from ranking. Candidate feature extraction is $O(PK)$, while sequential novelty requires $K(p-1)$ similarities at prompt $p$ and $KP(P-1)/2$ in total. Thus, for fixed $P$, the selector is linear in $K$, not quadratic within a prompt. Table~\ref{tab:complexity} confirms this trend on an RTX 4080 SUPER with $P=5$, 20 measured repeats, cached reference/text embeddings, and model initialization and diffusion generation excluded. DCAL adds no trainable parameters; generating the additional diffusion samples remains the dominant system-level cost.

\begin{table}[t]
\centering
\caption{Measured DCAL overhead versus candidate quantity. Times are milliseconds; peak memory is MiB.}
\label{tab:complexity}
\scriptsize
\setlength{\tabcolsep}{2.4pt}
\resizebox{\linewidth}{!}{%
\begin{tabular}{rrrrrr}
\toprule
$K$ & Images & Features & Ranking & Total & Peak mem. \\
\midrule
2 & 10 & 391.4 & 3.22 & 395.6 & 1430.7 \\
4 & 20 & 784.7 & 4.45 & 790.1 & 1945.0 \\
8 & 40 & 1582.1 & 7.56 & 1590.8 & 2969.8 \\
\bottomrule
\end{tabular}}
\end{table}

\begin{figure}[t]
\centering
\setlength{\tabcolsep}{1.2pt}
\begin{tabular}{@{}cccc@{}}
\scriptsize Reference & \scriptsize Random & \scriptsize Identity-only & \scriptsize DCAL \\
\includegraphics[width=0.205\linewidth]{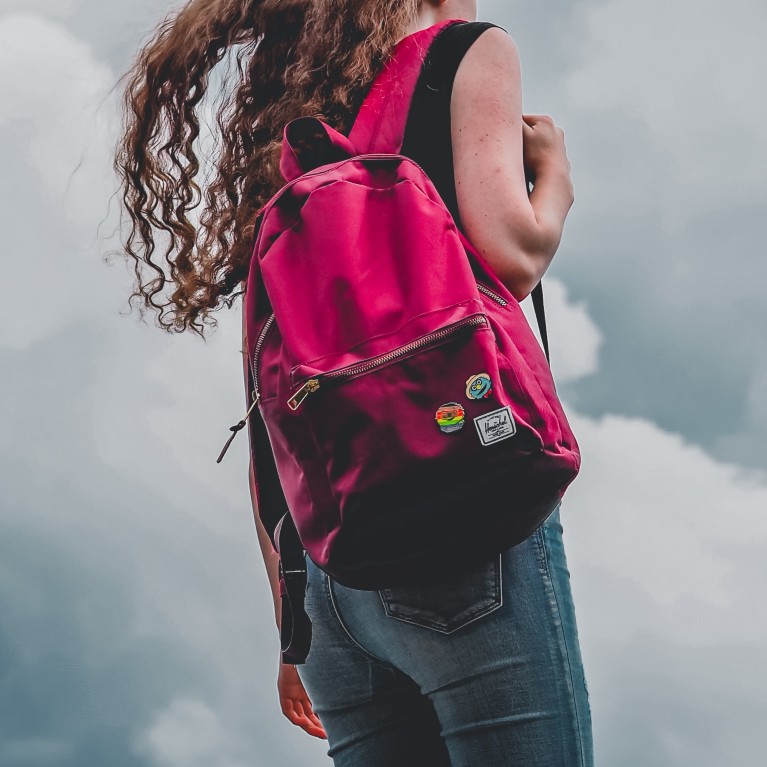} &
\includegraphics[width=0.205\linewidth]{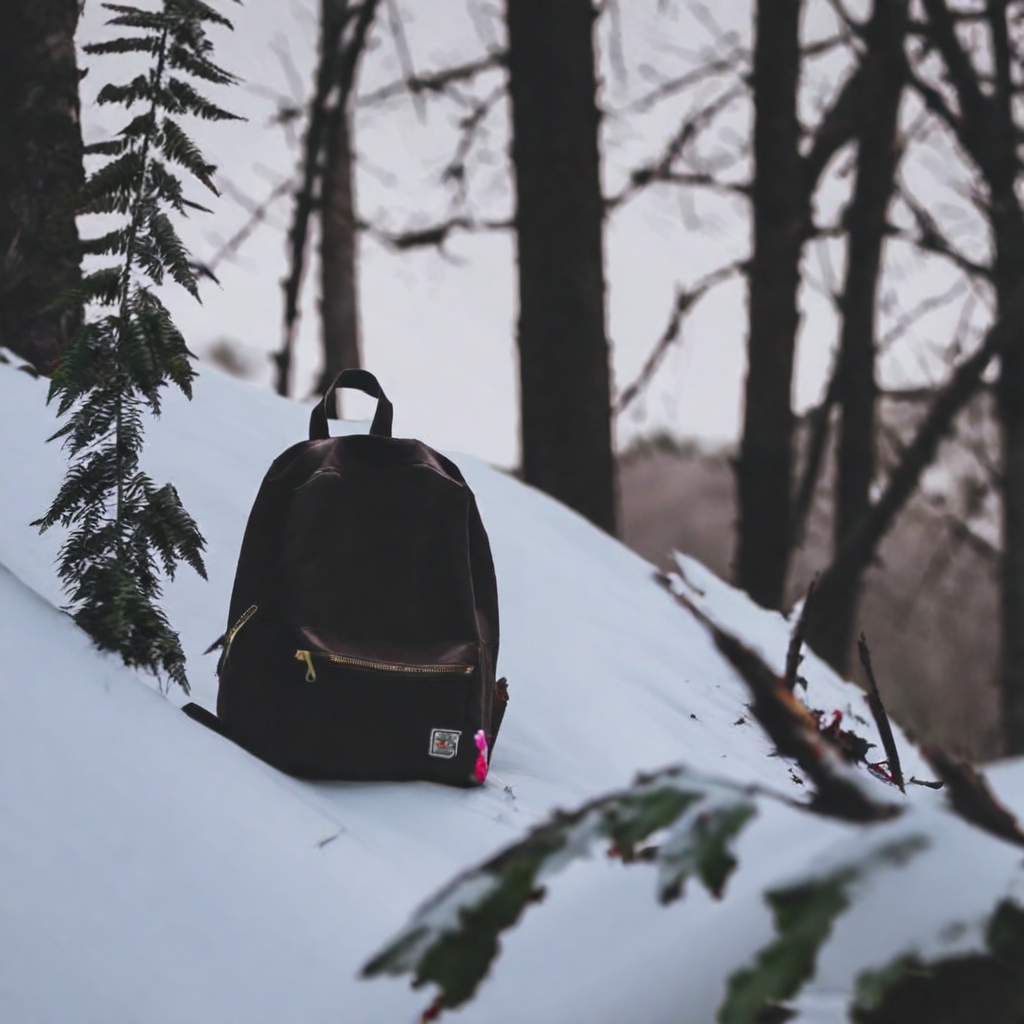} &
\includegraphics[width=0.205\linewidth]{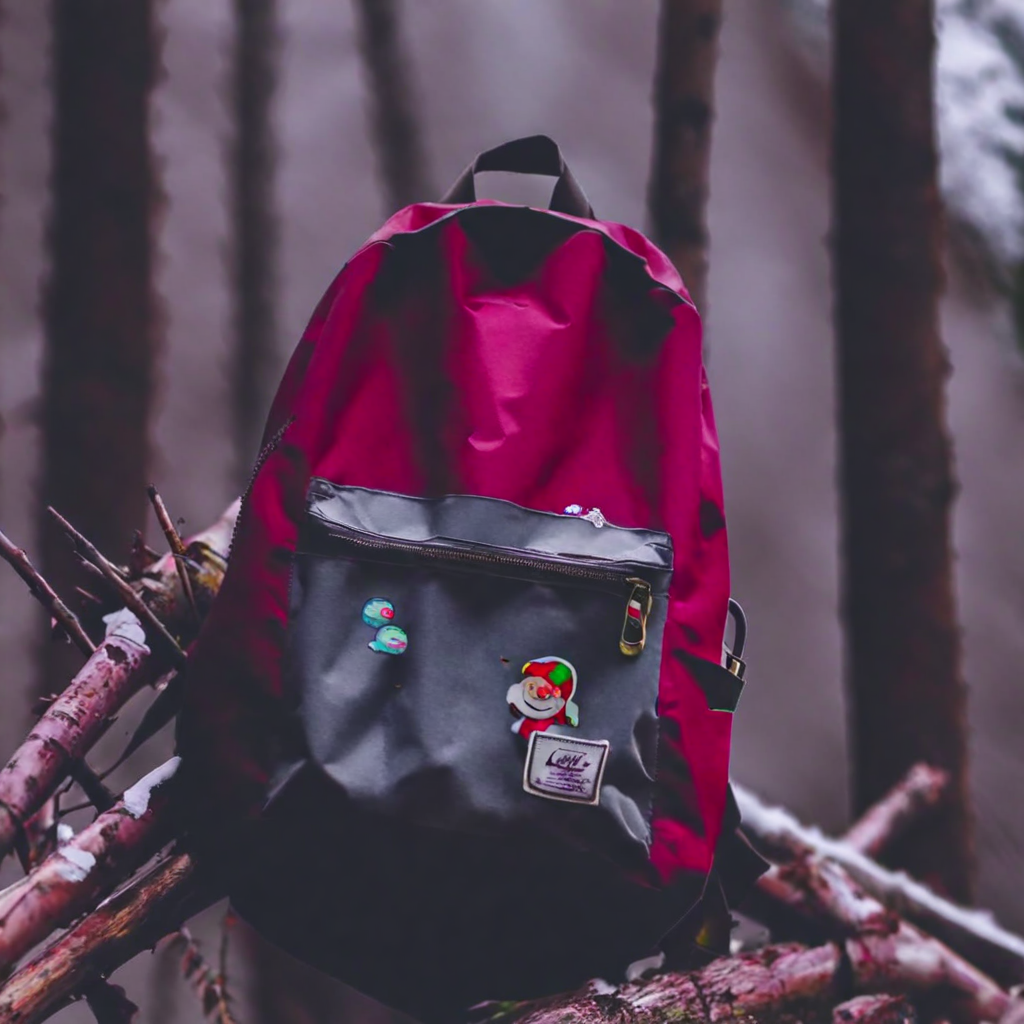} &
\includegraphics[width=0.205\linewidth]{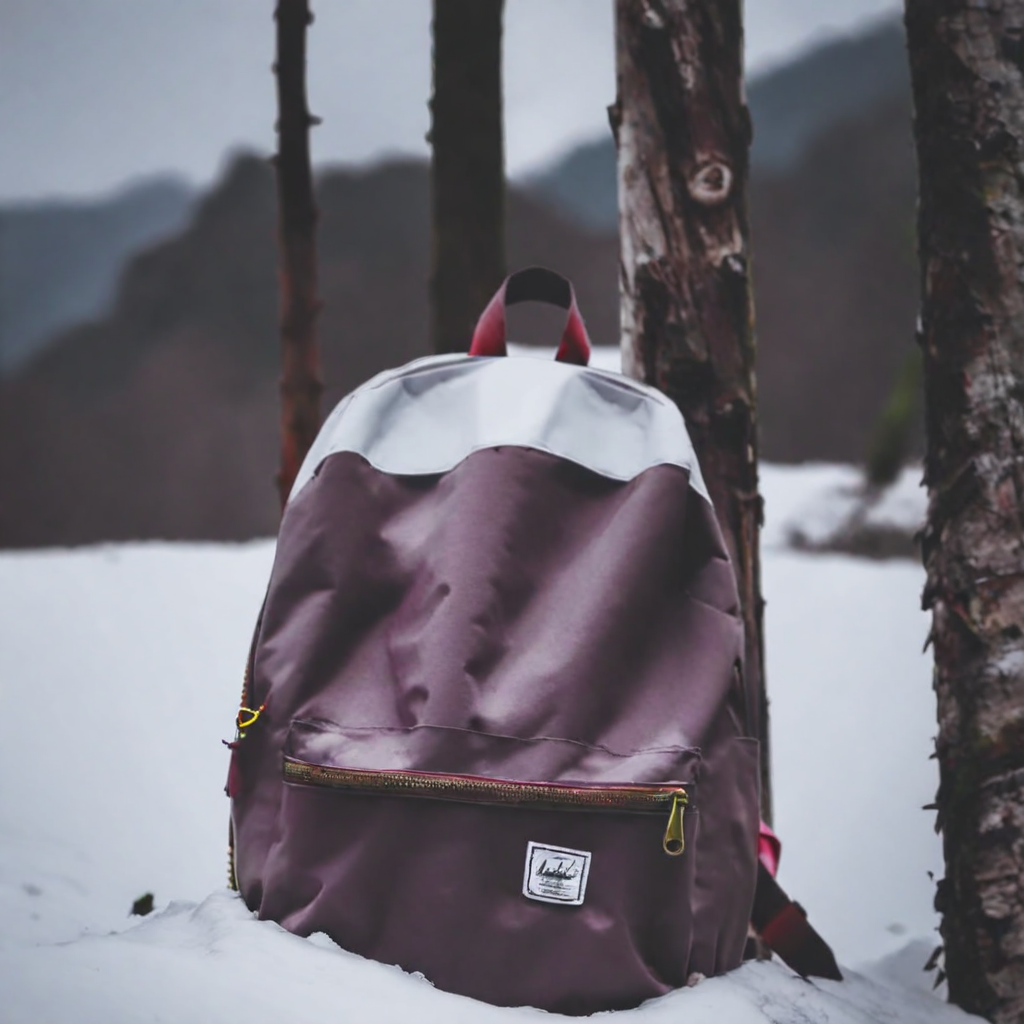} \\
\multicolumn{4}{c}{\scriptsize Backpack: ``a [V] backpack in the snow''} \\
\includegraphics[width=0.205\linewidth]{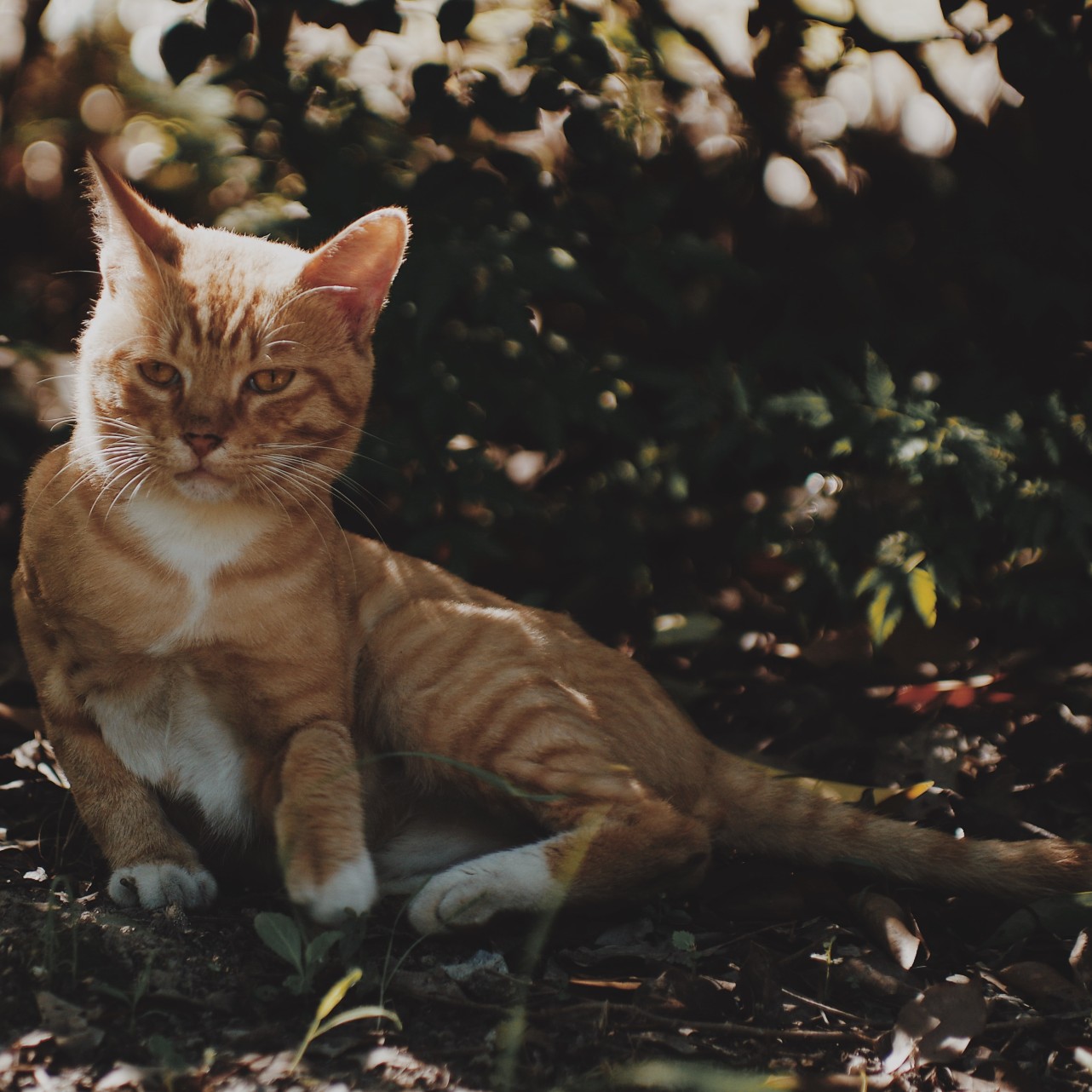} &
\includegraphics[width=0.205\linewidth]{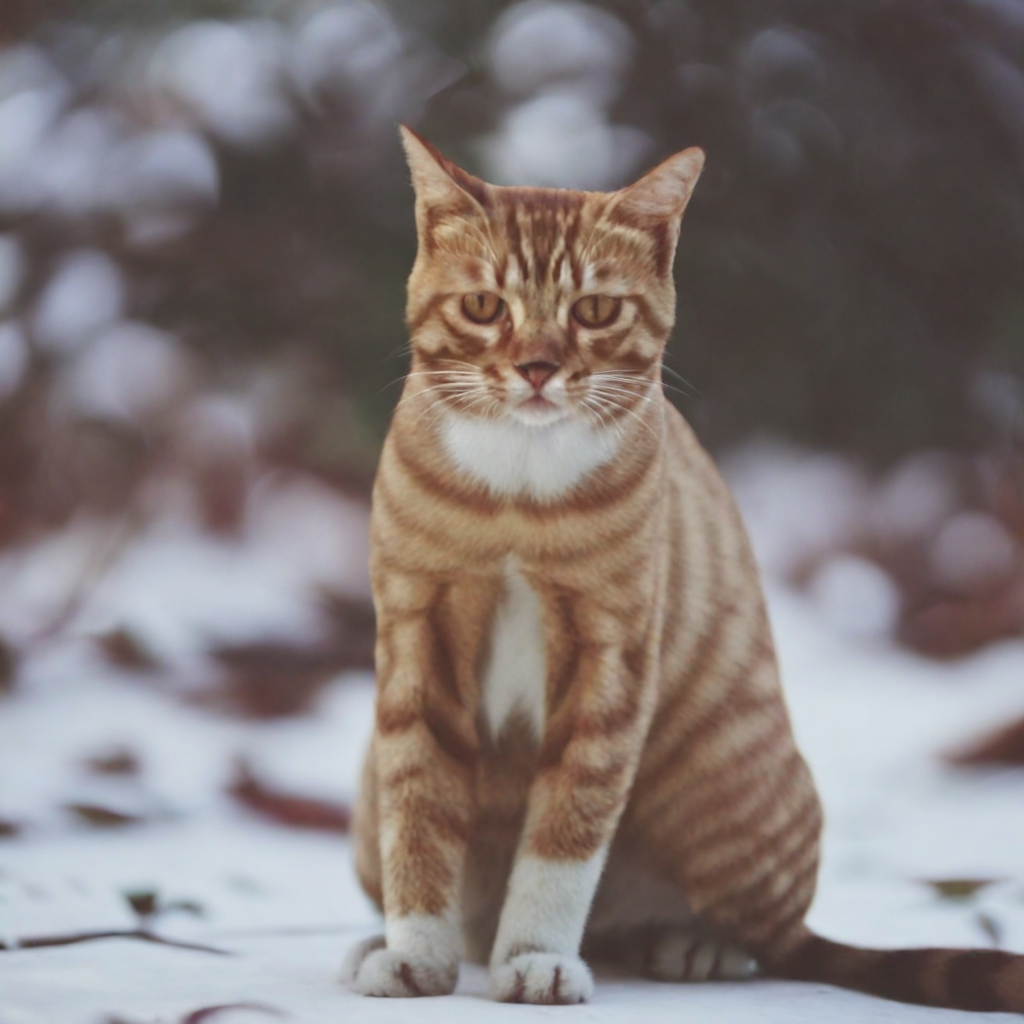} &
\includegraphics[width=0.205\linewidth]{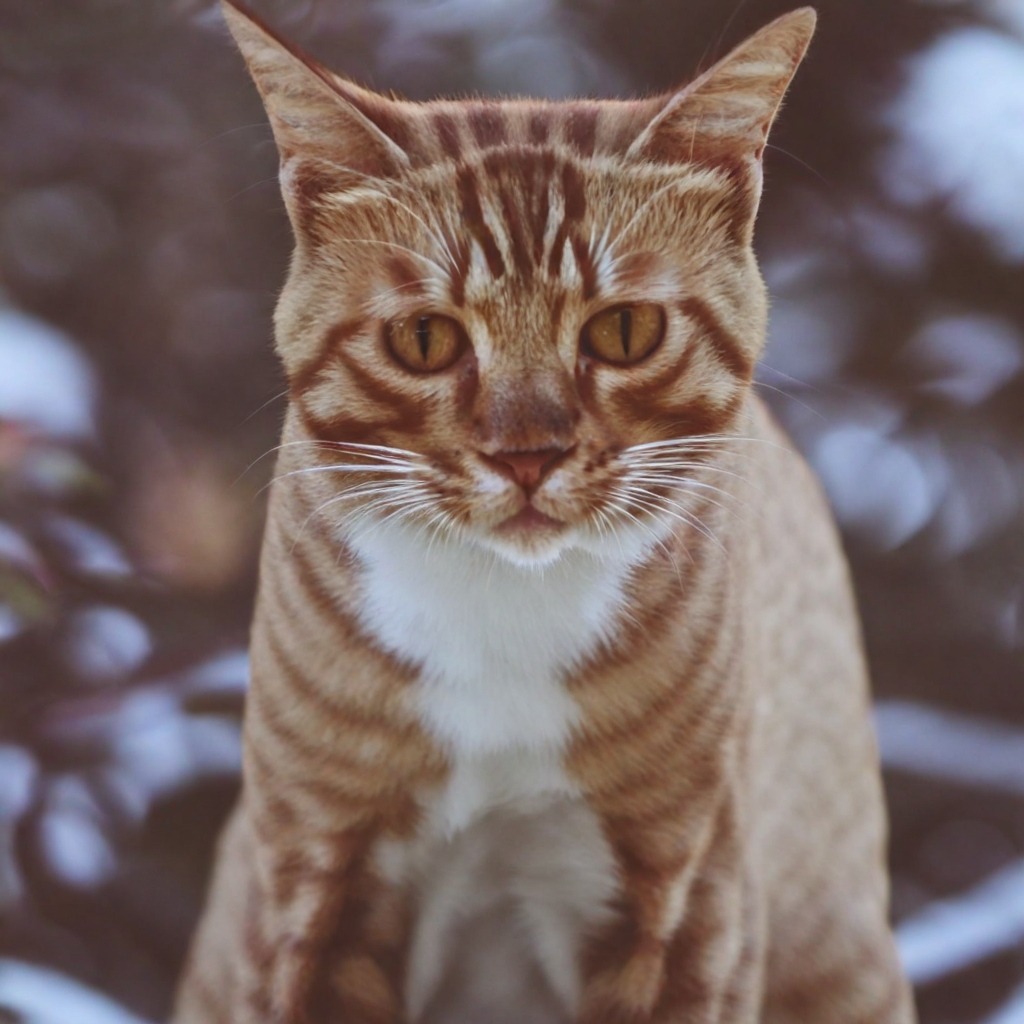} &
\includegraphics[width=0.205\linewidth]{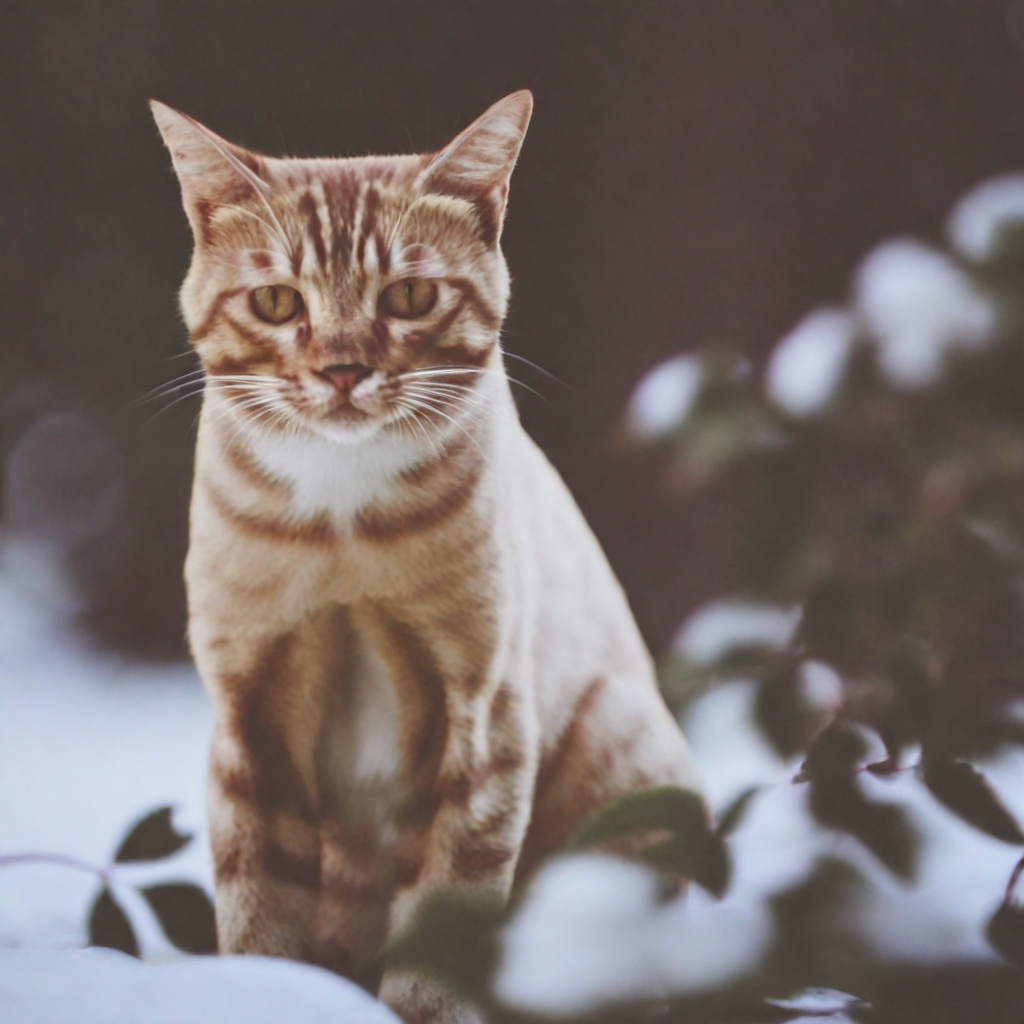} \\
\multicolumn{4}{c}{\scriptsize Cat: ``a [V] cat in the snow''}
\end{tabular}
\caption{Same-pool comparison with one training reference per subject. Each row reuses one $K=4$ pool; only the selector differs.}
\label{fig:qual}
\end{figure}

\subsection{Per-Subject Calibration Analysis}
Because several aggregate differences from identity-only selection are small, Table~\ref{tab:idtrade} also reports subject-level comparisons. After four exact ties, \DCAL{} achieves higher values for more subjects on CLIP-T and all three diversity metrics, whereas identity-only does so on CLIP-I and DINO-I. The resulting trade-off is distributed across subjects rather than concentrated in a small subset.

\begin{table}[t]
\centering
\caption{Subject-level comparison with identity-only-of-4. Entries count wins over the 30 subjects.}
\label{tab:idtrade}
\scriptsize
\setlength{\tabcolsep}{3.2pt}
\resizebox{\linewidth}{!}{%
\begin{tabular}{lcccccc}
\toprule
Selector & C-I & D-I & C-T & C-Div & D-Div & Pair-LP \\
\midrule
\DCAL{} & 7 & 3 & \textbf{25} & \textbf{17} & \textbf{21} & \textbf{16} \\
Identity-only & \textbf{19} & \textbf{23} & 1 & 9 & 5 & 10 \\
Ties & 4 & 4 & 4 & 4 & 4 & 4 \\
\bottomrule
\end{tabular}}
\end{table}

Prompt-level decisions remain auditable because candidate identifiers and component scores are saved before ranking. Changing the weights therefore reorders a fixed pool without another sampling or adapter-training run, while preserving the exact candidate and scores behind each returned image.

\subsection{Interpretation of the Diversity Score}
The novelty term in Eq.~\eqref{eq:novelty} rewards candidates that are less redundant with earlier selections for the same subject. It neither measures average distance to the other candidates of the current prompt nor optimizes diversity across subjects. Subject-pool standardization prevents CLIP-I, DINO-I, or CLIP-T scale from dominating, while prompt-wise novelty standardization prevents its smaller raw cosine-distance range from making $\lambda_D$ ineffective. The weights express a fixed preference among calibrated identity, text, and cross-prompt novelty, not universal calibration over the generator distribution.

Two limiting cases clarify the formulation: $\lambda_D=0$ retains identity and text only, while $\lambda_T=\lambda_D=0$ gives standardized identity and differs from the absolute-score identity-only control in Sec.~III-C. The reported 30-subject evaluation setting is fixed rather than selected per subject.

Accordingly, DCAL should be interpreted as a calibrated multi-objective selector rather than an optimizer of any individual identity or diversity metric.

\section{Discussion}
\textbf{Relation to existing re-ranking methods.} DCAL complements rather than replaces Pareto sorting or learning-to-rank. For fixed generators and candidate pools, it combines calibrated identity, text, and cross-prompt redundancy without retraining while controlling for candidates, seeds, and generators.

\textbf{Role of same-pool controls.} The fixed-seed Random-of-4 control provides a same-count reference, while identity-only-of-4 and \DCAL{} compare objectives on identical images, isolating ranking effects.

\textbf{Deployment considerations.} \DCAL{} permits weight changes without retraining but incurs the cost of $K$ diffusion outputs and CLIP/DINOv2 feature extraction. It is intended for settings in which selection quality is prioritized over inference throughput.

\textbf{Limitations.} \DCAL{} requires $K$ candidates and therefore costs more than single-sample generation. It is a predefined multi-prompt sequence selector rather than an order-invariant single-prompt rule. All experiments use the fixed benchmark prompt order; permuting that order may change later novelty scores and selections, and a single prompt reduces to identity-text scoring. The current evaluation uses the public DreamBooth 30-subject protocol and one DreamBooth LoRA $r=16$ pool. Broader cross-dataset evaluation and supervised learned-ranker baselines remain future work. Transfer to other personalization backbones may require recalibrating the score standardization and weights.

\section{Conclusion}
We presented \DCAL{}, a training-free sequential selector for personalized text-to-image generation. Under identical candidate counts and pre-generated candidate pools, \DCAL{} achieves higher 1-LPIPS, CLIP-I, DINO-I, and CLIP-T than the fixed-seed Random-of-4 control, while comparison with identity-only selection quantifies the trade-off among identity preservation, prompt alignment, and diversity. The method is auditable, requires no adapter retraining, and selects one image for each prompt in a predefined prompt set.

\end{document}